\documentclass{article}

\usepackage[preprint,nonatbib]{neurips_2024}

\usepackage[utf8]{inputenc} 
\usepackage[T1]{fontenc}    
\usepackage{hyperref}       
\usepackage{url}            
\usepackage{booktabs}       
\usepackage{amsfonts}       
\usepackage{nicefrac}       
\usepackage{microtype}      
\usepackage{xcolor}         

\usepackage{amsmath}
\usepackage{tcolorbox}
\usepackage{graphicx}
\usepackage{subcaption}
\usepackage{wrapfig}
\usepackage{lipsum}  

\title{
Refining Time Series Anomaly Detectors using Large Language Models
}

\author{%
Alan Yang$^1$\thanks{This work was completed during an internship at Meta.},\;
Yulin Chen$^2$,\;
Sean Lee$^2$,\;
Venus Montes$^2$
\\
$^1$Stanford University,\;$^2$Meta 
\\
\texttt{yalan@stanford.edu},\;
\texttt{\{yulinchen,seunghak,vmd\}@meta.com} \\
}

\begin{document}

\maketitle

\begin{abstract}
Time series anomaly detection (TSAD) is of widespread interest across many industries, including finance, healthcare, and manufacturing. Despite the development of numerous automatic methods for detecting anomalies, human oversight remains necessary to review and act upon detected anomalies, as well as verify their accuracy. We study the use of multimodal large language models (LLMs) to partially automate this process. We find that LLMs can effectively identify false alarms by integrating visual inspection of time series plots with text descriptions of the data-generating process. By leveraging the capabilities of LLMs, we aim to reduce the reliance on human effort required to maintain a TSAD system.
\end{abstract}

\section{Introduction}

Time series anomaly detection (TSAD) is a long-standing research area that has seen significant advancements in recent years. Despite the development of various methods, including those based on quickest change detection \cite{bayart2005walter,page1954continuous}, machine learning \cite{ramaswamy2000efficient,nakamura2020merlin}, and deep learning approaches \cite{xu2022transformer}, TSAD often requires substantial human effort to implement and maintain effectively. It generally requires a human decision maker to determine whether an detected anomaly is the result of noise, changing operating conditions, or a genuine issue that requires intervention. In this work, we explore the application of using LLMs to solve the zero-shot classification problem of determining whether a potential anomaly found by a detector is a true positive or false positive. By harnessing their ability to compare a potentially anomalous time series against predicted values and reason with textual context surrounding the data-generating process, LLMs have the potential to significantly reduce the human effort required for processing false alarms produced by TSAD systems.

Recent studies have demonstrated the potential of large language models (LLMs), particularly multimodal LLMs, in time series tasks such as pattern recognition, reasoning, and anomaly detection \cite{cai2024timeseriesexam,xie2024chatts,zhuang2024see,zhou2024can,alnegheimish2024large}. Current LLM-based time series anomaly detection (TSAD) methods can be broadly categorized into two approaches: 1) prompting an LLM with time series data to identify anomalies and their locations \cite{cai2024timeseriesexam,zhuang2024see,alnegheimish2024large}, or 2) utilizing the LLM as a time series forecaster and comparing the forecast against actual values \cite{jin2023time,sarfrazposition,alnegheimish2024large}. The advantages of these methods lie in their ability to perform reasonably well with minimal training data and incorporate textual context about the data-generating process into the detection process. Furthermore, the capabilities of LLMs are expected to improve as they are trained on more diverse datasets, including visualizations such as charts and graphs \cite{hendrycks2020measuring}.

Despite their potential advantages, LLMs incur high inference costs for long-running time series with high volumes of data. Moreover, their performance can be inconsistent, and they often underperform against simple, more traditional methods \cite{merrill2024language,tan2024language,sarfrazposition}.
To address these limitations, we propose a novel approach that combines the strengths of both traditional and LLM-based TSAD methods.
By utilizing LLMs to evaluate potential anomalies discovered by an existing detector, rather than relying on them to identify anomalies independently, our method can refine the accuracy of traditional detectors while minimizing LLM inference costs.

Our proposed approach offers several key benefits. Firstly, LLMs can effectively filter out false alarms, thereby improving detector precision without compromising recall. Additionally, the method provides explanations for its decisions, like other LLM-based TSAD methods \cite{zhuang2024see}. Furthermore, it can incorporate contextual information specific to the data-generating process, which can be challenging to encode into non-LLM-based TSAD methods. The operational cost is also reduced, as LLM inference is only required to evaluate a small fraction of the time series segments identified by the existing anomaly detector. Finally, the task of verifying the correctness of an existing anomaly detector is a more straightforward task that requires less reasoning ability than asking an LLM to identify all anomalous regions in a time series or forecast future values \cite{alnegheimish2024large}.

We demonstrate the approach by refining a simple detector based on $k$-nearest neighbor ($k$-NN) search \cite{ramaswamy2000efficient} using LLMs from the multimodal \texttt{Llama3.2} family \cite{dubey2024llama}. Using a collection of datasets in the Hexagon ML/UCR time series anomaly detection archive \cite{wu2021current}, we show that LLMs can improve the best-possible $k$-NN detectors' precision and recall, demonstrating their ability to distinguish true anomalies from false positives.

\section{Time series anomaly detection}\label{s-tsad}

In this work, we consider time series data $X=x_1,x_2,\ldots,x_{T}$,
where $x_t\in\mathbb{R}$ is the value of the time series at time $t$, $t\in[1,T]$ is the time step or epoch, and $T$ is the time series length. We represent a time series anomaly using a contiguous time interval $A\subset[1,T]$, and the set of anomalous intervals by $\mathcal A$. A \emph{time series anomaly detector} accepts $X$ and returns a set of contiguous time intervals $\mathcal I$, such that each interval $I\in\mathcal I$ represents a potential anomaly $I\subseteq [1,T]$. The goal of a detector is to find a set $\mathcal I$ that covers as little of $[1,T]$ as possible, while covering as much of the anomalies $\cup\mathcal A$ as possible. An interval $I$ is considered a true positive if $|A\cap I|>0$ for some $A\in\mathcal A$, \emph{i.e.}, it overlaps with a true anomalous interval. Otherwise, $I$ is considered to be a false positive.

\subsection{Anomaly detection using nearest-neighbor search}\label{s-knn}

In this subsection, we present an anomaly detector based on $k$-NN search \cite{ramaswamy2000efficient}. Although simple, it often works as well, if not better than, more sophisticated methods in practice \cite{goswami2022unsupervised,sarfrazposition}.
The $k$-NN detector starts with a historical time series $X'=x'_1,x'_2,\ldots,x'_{N}$ collected prior $X$, which is assumed to contain no anomalies. The historical data $X'$ is then converted into $N-h$ sliding intervals of length $h$:
\[
w'_i = x'_i,x'_{i+1},\ldots,x'_{i+h}, \quad i=1,\ldots,N-h.
\]
At test time, the $k$-NN detector considers length-$h$ sliding intervals of $X$. Let the window starting at time $t\in[1,T]$ be $w_t = x_t,x_{t+1},\ldots,x_{t+h}$,
and let $\hat w_t$ be the $k$-th nearest neighbor to $w_t$ among the historical windows $w'_1,\ldots,w'_{N-h}$ as measured by Euclidean distance. Then, $w_t$ is considered to contain an anomaly if $\|w_t - \hat w_t\|_2 \ge \tau$, where $\tau$ is a threshold parameter. The $k$-NN detector outputs a set $\mathcal I$ of length-$h$ intervals $[t,t+h]$ for which the threshold was exceeded.




\section{Proposed method}\label{s-proposed}

\begin{figure}
\centering
\includegraphics[width=\textwidth]{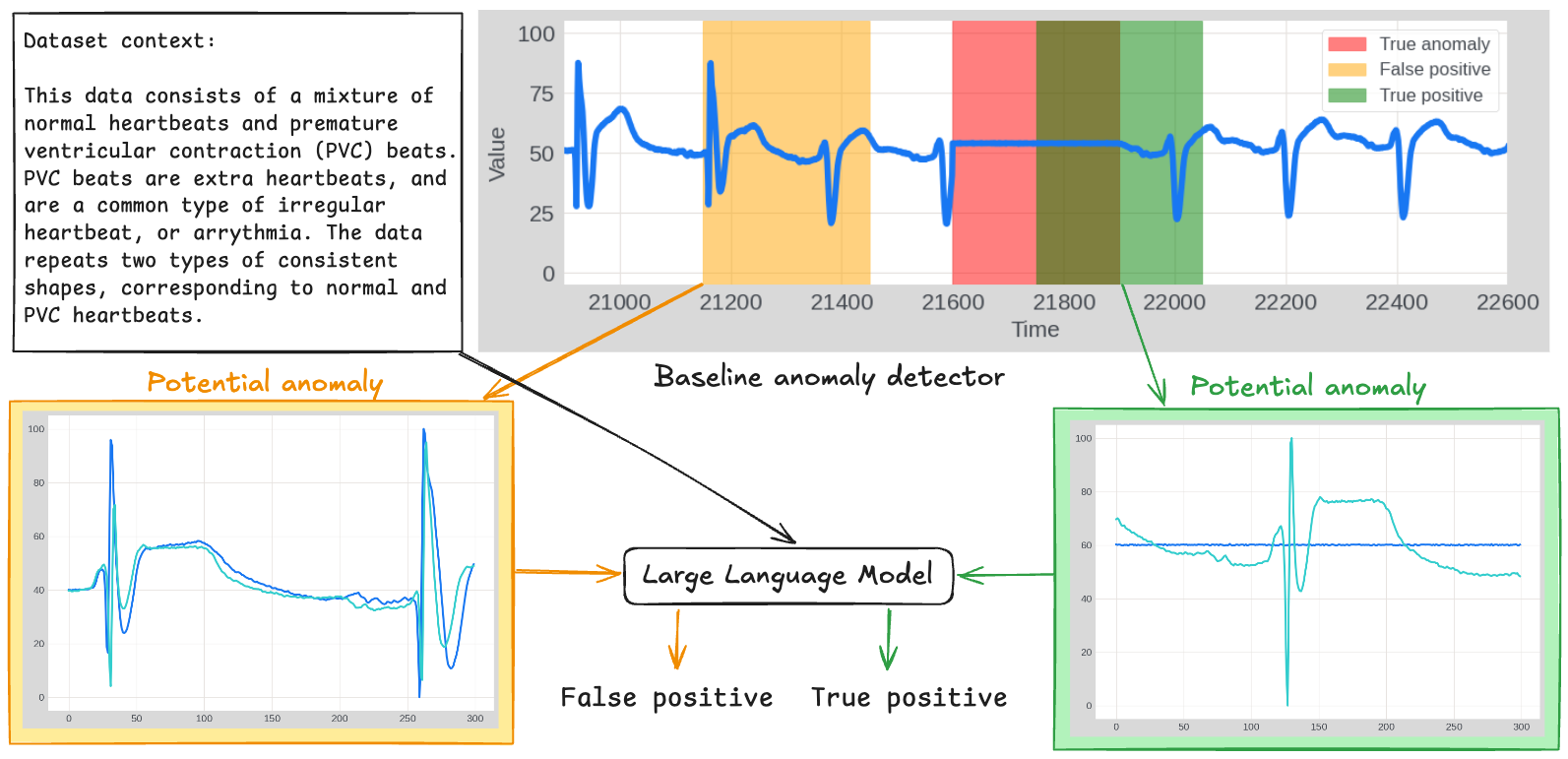}
\caption{Proposed approach, illustrated using a heartbeat dataset \cite{wu2021current}. Potential anomalies are first identified by a baseline anomaly detector. An LLM then determines whether they are true or false positives by comparing the time series against a forecast/historical data, taking into account a text description of the context around the data-generating process.}
\label{f-system-diagram}
\end{figure}

In this work, we propose an approach to enhance the performance of existing time series anomaly detectors by leveraging the capabilities of large language models (LLMs). Our method involves using an LLM as a post-processing step for the outputs of a baseline anomaly detector, which can be any type of detector, ranging from simple threshold-based detectors, such as the $k$-NN detector presented in Section \ref{s-knn}, to more complex ones based on sophisticated forecast models.

The proposed approach consists of two stages. Initially, the baseline anomaly detector identifies a set of potentially anomalous intervals $\tilde{\mathcal I}$. Subsequently, for each interval $I\in\tilde{\mathcal I}$, we prompt the LLM to determine whether it is a true positive or false positive.
If the LLM classifies $I$ as a true positive, it is included in the final set of predicted anomalies $\mathcal I$. Conversely, if the LLM determines that $I$ is a false positive, it is excluded from the final set.
The resulting set $\mathcal I\subseteq\tilde{\mathcal I}$ is a refined subset of the baseline detector's predictions. Since the LLM does not detect new anomalies, our approach cannot increase the recall of the baseline anomaly detector, but it can improve the detector's precision by filtering out false alarms.

To help the LLM make a decision about each potentially anomalous interval $I\in\tilde{\mathcal I}$, we provide a plot of the time series $X$ over the interval $I$ in blue, overlaid on a plot of the predicted values of that time series on the same interval in green. Those predicted values can come from a forecasting model of the time series, or it can come from historical values of the time series data. In the case of the $k$-NN detector, the $k$-th nearest neighbor interval may be used as the prediction. Besides the plot, a text description that provides context on the data generating process and expected behavior of the time series is also provided to the LLM. The proposed method is illustrated in Figure \ref{f-system-diagram}.

We used the following prompt format:

{\footnotesize
\texttt{``Does the blue time series have the same shape as the green time series?
First answer the question focusing on the beginning of the time series, then the middle, then finally the end.}
\texttt{If the answer is Yes, then the blue time series should also match the following description: \emph{(dataset context)}}

\texttt{The time series data are plotted in the given images. Use visual inspection to draw your conclusions. Consider the shapes of the plotted time series.''}}

\section{Experiments}\label{s-experiments}

\paragraph{Datasets.}
In our experiments, we use 18 datasets from the Hexagon ML/UCR time series anomaly detection archive \cite{wu2021current}, which contains a collection of diverse univariate time-series of varying lengths spanning domains such as medicine, sports, entomology, and space science, with both natural and synthetic anomalies. Each dataset consists of an anomaly-free \emph{training} time series, and a \emph{test} time series that contains a single anomalous interval. Additional details can be found in Appendix \ref{a-data}.

\paragraph{Baseline anomaly detector.}
We use the $k$-NN detector described in Subsection \ref{s-knn} as our baseline anomaly detector to test our approach. Since our focus is on assessing the LLM rather than the baseline anomaly detector itself, we chose the detector parameters to be near-optimal, to determine whether the LLM can improve upon a good detector. We chose the window size $h$ to be $10\%$ larger than the length of the true anomaly interval for each dataset, up to minimum and maximum values $h_{\min}=30$ and $h_{\max}=300$. We chose $\tau$ to be $90\%$ of the largest threshold capable of detecting the anomaly in the test data. For simplicity, we we only use every $\lfloor h/3\rfloor$-th window. For example, if $h=100$, we only consider windows $w_1, w_{50}, w_{100},\ldots$, \emph{etc.} The above choices of parameters approximate an expertly-tuned baseline anomaly detector, and ensure that the potential anomalies identified are sufficiently non-trivial. Finally, we used $k=3$ neighbors for all datasets.

\paragraph{LLM.}
In our experiments, we used the vision-capable \texttt{llama3.2-90b-vision-instruct} and \texttt{llama3.2-11b-vision-instruct} multimodal LLMs \cite{dubey2024llama}. For those models, we also compared against the case where the dataset context is ablated, \emph{i.e.}, we only ask the model whether or not the shape of the data matches that of the prediction, without providing the textual dataset context. To make the model outputs more consistent, we run each model inference five times, and take the majority vote. Therefore, an interval is declared a true positive if the model considered it to be anomalous three out of the five times. We also considered a version of the experiments where we used the text-only \texttt{llama3.3-70b-instruct}, and passed the time series values as plain text values. Details about the text-only case are given in Appendix \ref{a-text-only}.

\paragraph{Metrics.}
The $k$-NN baseline anomaly detector produced $168$ potentially anomalous intervals across $18$ datasets, $39$ of which are true positives. We evaluate the LLM by the reduction in false positive intervals and the percentage of true positive intervals retained. We also consider the final number of anomalies detected. That is, we report the number of datasets for which at least one true positive interval was correctly recognized by the model. This third metric is not the same as the percentage of true positive intervals retained, since several datasets had more than one true positive detected interval. We chose the above metrics since we are primarily interested in reducing the number of false alarms that require human review without compromising detection power.

\subsection{Results}\label{s-results}

The results are summarized in Table \ref{t-performance}. Each row gives the performance of different LLM model choices, with and without the dataset context provided in the prompt. Different choices of models lead to different performance trade-offs. The 90B parameter vision model without dataset context had the greatest reduction in false alarms, but missed almost a fifth of the anomalies. On the other hand, the 70B-parameter text-only model without dataset context classified every interval as a true positive.
Examples of the time series data, true and false positive intervals, and the corresponding model responses are given in Appendix \ref{a-add-results}.

\begin{table}[b]
\caption{Performance comparison across 18 datasets, including the amount of false positive (FP) intervals reduced, true positive (TP) intervals retained, and anomalies detected.}
\label{t-performance}
\centering
\begin{tabular}{llll}
Model     & FPs reduced & TPs retained & Anomalies detected \\
\midrule
\texttt{llama3.2-vision-90b} & 26\%  & 85\%  & 94\% \\
\texttt{llama3.2-vision-11b} & 48\% & 85\%  & 89\%  \\
\texttt{llama3.3-70b} & 5.4\% & 97\% & 94\% \\
\texttt{llama3.2-vision-90b} (no dataset context) & 58\% & 67\% & 78\% \\
\texttt{llama3.2-vision-11b} (no dataset context) & 5.4\% & 100\% & 100\% \\
\texttt{llama3.3-70b} (no dataset context) & 0\% & 100\% & 100\% \\
\bottomrule
\end{tabular}
\end{table}

\section{Conclusions}\label{s-conclusions}

In this paper, we investigated the use of multimodal LLMs to identify false alarms predicted by a TSAD system. Our results show that the \texttt{llama3.2-vision} models can effectively distinguish false alarms from true positives in real-world data. This approach not only has the potential to reduce human effort in reviewing potential anomalies but also offers a promising solution for automatically labeling time series anomalies, which is crucial for training and tuning effective detectors.






\bibliographystyle{IEEEtran}
\bibliography{tsad_llm}


\appendix

\section{Data}\label{a-data}

Table \ref{t-datasets} contains some details about the 18 datasets from the Hexagon ML/UCR time series anomaly detection archive \cite{wu2021current}. The table also gives the number of true positive and false positive intervals detected by the baseline $k$-NN anomaly detector.

\section{Text-only model experiments} \label{a-text-only}

We also compared against the text-only \texttt{llama3.3-70b-instruct} model. In this case, the last three sentences of the prompt were as follows:

{\footnotesize
\texttt{``Does the blue time series have the same shape as the green time series?
First answer the question focusing on the beginning of the time series, then the middle, then finally the end.}
\texttt{If the answer is Yes, then the blue time series should also match the following description: \emph{(dataset context)}}

\texttt{In the time series data given below, each step is separated by a comma. In your analysis, try not to repeat large chunk of values in the time series to save space.''}

\texttt{(time series values)}}

The time series values are provided using in a three-column format. The first column contains the time index, the second contains the values of the potentially anomalous blue time series, and the third column contains the values of the predicted green time series. The values of the time series are first scaled and shifted to be between the values $[0,1]$, and then quantized to two significant figures.

\section{Additional results}\label{a-add-results}

\begin{figure}
\centering
\begin{subfigure}[b]{\textwidth}
\includegraphics[width=\textwidth]{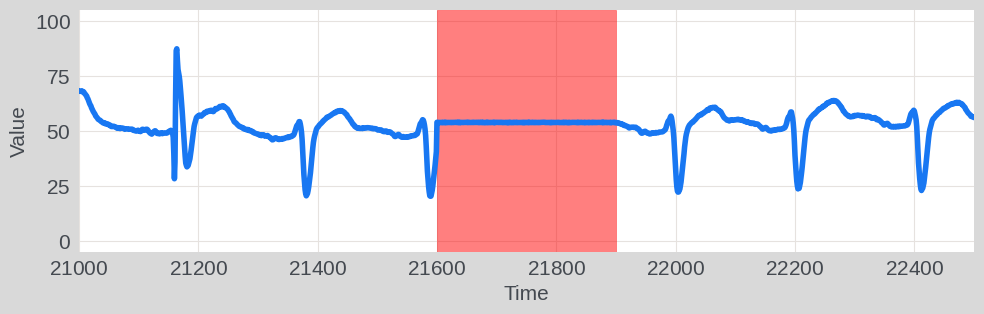}
\caption{Time series values}
\end{subfigure}
\begin{subfigure}[b]{0.48\textwidth}
\includegraphics[width=\textwidth]{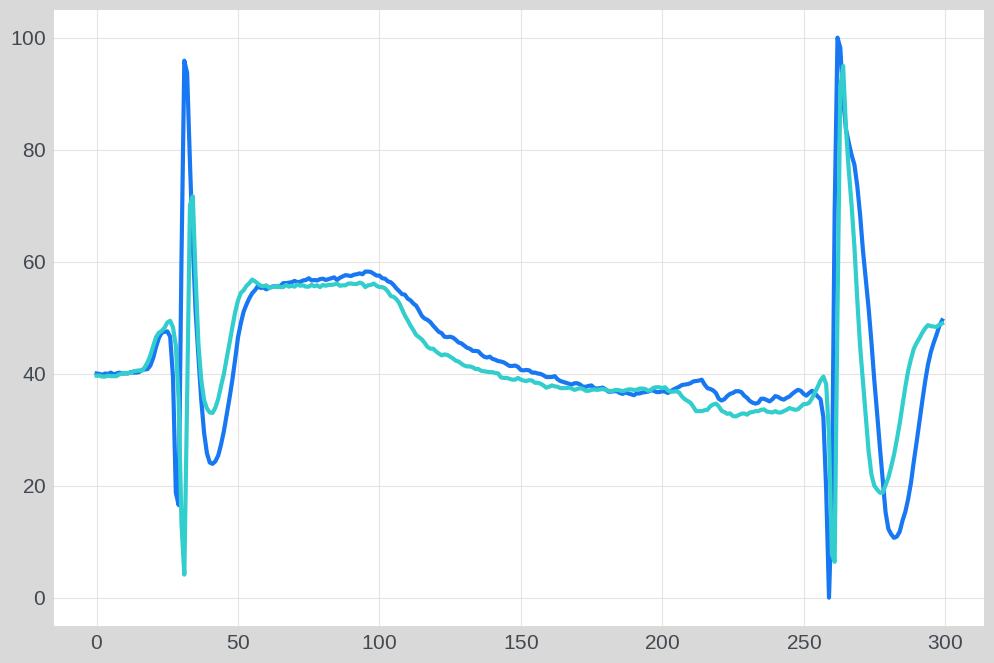}
\caption{False positive example}
\end{subfigure}
\hfill
\begin{subfigure}[b]{0.48\textwidth}
\includegraphics[width=\textwidth]{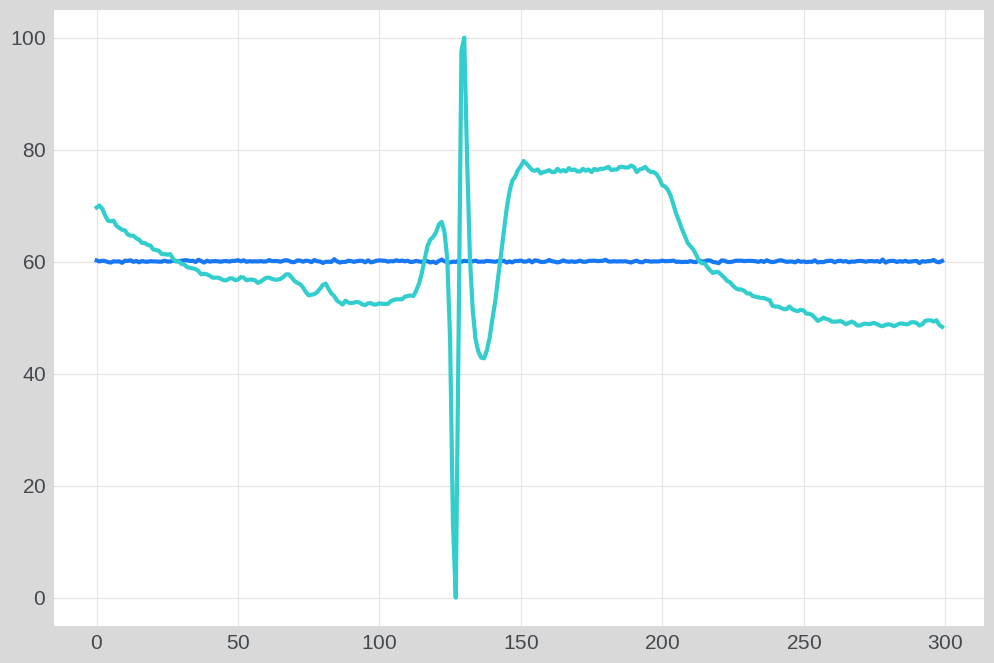}
\caption{True positive example}
\end{subfigure}
\begin{subfigure}[b]{\textwidth}
\vspace{1em}
\begin{minipage}[t]{\textwidth}
\scriptsize
\texttt{
**Step 1: Analyze the beginning of the time series**
The blue and green time series start at approximately the same value, around 40. The initial shape of both time series appears to be similar, with a slight increase followed by a decrease.
**Step 2: Analyze the middle of the time series**
In the middle section of the time series, both the blue and green lines exhibit a similar pattern of fluctuations. They both have peaks and troughs that seem to mirror each other, although the exact values may differ slightly.
**Step 3: Analyze the end of the time series**
Towards the end of the time series, the blue and green lines again show a similar trend. They both decrease and then increase, with the blue line ending at a slightly higher value than the green line.
**Conclusion**
Based on visual inspection, it appears that the blue time series has a similar shape to the green time series throughout its entire length. The shapes of the plotted time series are consistent with the description provided, which mentions a mixture of normal heartbeats and premature ventricular contraction (PVC) beats. Therefore, the answer is:
**Yes**, the blue time series matches the description of consisting of a mixture of normal heartbeats and PVC beats.}
\end{minipage}
\caption{False positive model response (correct)}
\end{subfigure}
\hfill
\begin{subfigure}[b]{\textwidth}
\vspace{1em}
\begin{minipage}[t]{\textwidth}
\scriptsize
\texttt{The blue time series does not have the same shape as the green time series. 
At the beginning of the time series, the blue line is flat, while the green line has a peak. In the middle, the blue line remains flat, but the green line has another peak. At the end, the blue line is still flat, while the green line has a dip.
Therefore, the blue time series does not match the description of the data consisting of a mixture of normal heartbeats and premature ventricular contraction (PVC) beats.}
\end{minipage}
\caption{True positive model response (correct)}
\end{subfigure}
\caption{Model responses for the \emph{2sddb40} dataset, which has dataset context: \emph{This data consists of a mixture of normal heartbeats and premature ventricular contraction (PVC) beats. PVC beats are extra heartbeats, and are a common type of irregular heartbeat, or arrythmia. The data repeats two types of consistent shapes, corresponding to normal and PVC heartbeats.}}
\label{f-sddb40}
\end{figure}

\begin{figure}
\centering
\begin{subfigure}[b]{\textwidth}
\includegraphics[width=\textwidth]{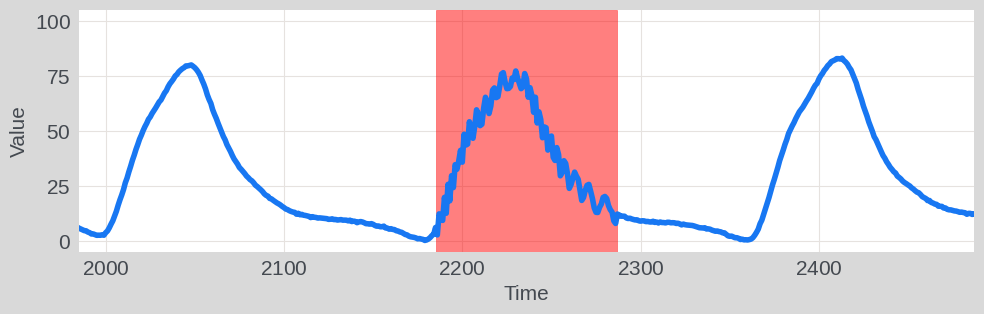}
\caption{Time series values}
\end{subfigure}
\begin{subfigure}[b]{0.48\textwidth}
\includegraphics[width=\textwidth]{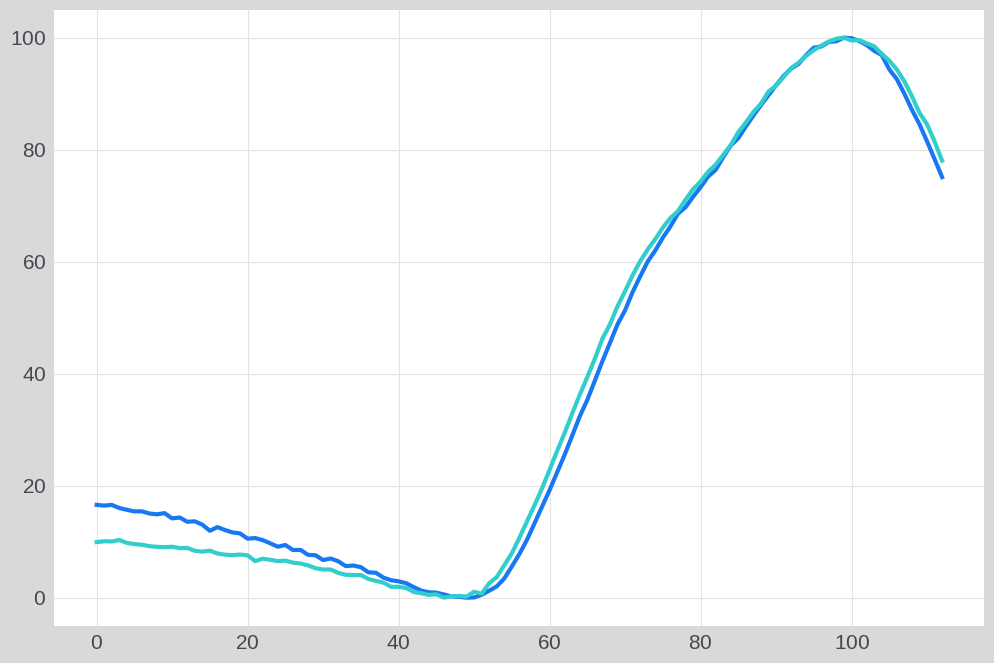}
\caption{False positive example}
\end{subfigure}
\hfill
\begin{subfigure}[b]{0.48\textwidth}
\includegraphics[width=\textwidth]{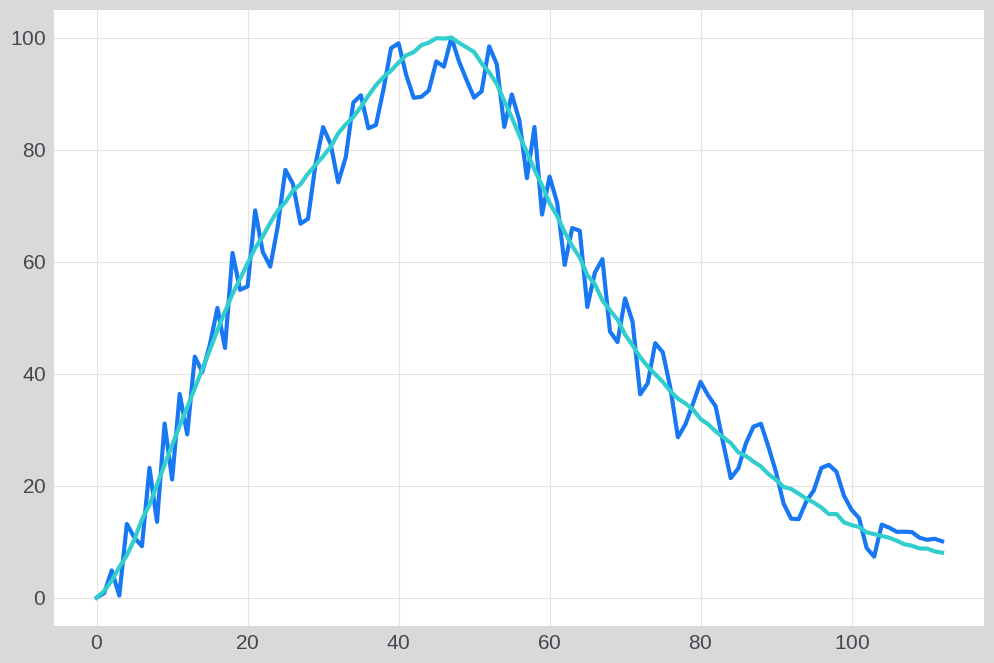}
\caption{True positive example}
\end{subfigure}
\begin{subfigure}[b]{\textwidth}
\vspace{1em}
\begin{minipage}[t]{\textwidth}
\scriptsize
\texttt{The blue time series and the green time series do not have the same shape.
**Beginning of the Time Series**
*   The blue time series starts at a higher value than the green time series.
*   The blue time series has a more gradual increase in the beginning, while the green time series has a steeper increase.
**Middle of the Time Series**
*   The blue time series has a more pronounced peak in the middle, while the green time series has a flatter peak.
*   The blue time series has a more rapid decline after the peak, while the green time series has a more gradual decline.
**End of the Time Series**
*   The blue time series ends at a lower value than the green time series.
*   The blue time series has a more rapid decrease at the end, while the green time series has a more gradual decrease.
**Conclusion**
Based on the visual inspection of the time series data, it is clear that the blue time series and the green time series do not have the same shape. The blue time series has a more pronounced peak, a more rapid decline, and a lower ending value compared to the green time series. Therefore, the blue time series does not match the description of the data containing arterial blood pressure measurements of a pig.}
\end{minipage}
\caption{False positive model response (incorrect)}
\end{subfigure}
\hfill
\begin{subfigure}[b]{\textwidth}
\vspace{1em}
\begin{minipage}[t]{\textwidth}
\scriptsize
\texttt{**Step 1: Analyze the beginning of the time series**
The blue time series starts at a low value and increases rapidly, while the green time series starts at a higher value and increases more gradually. The shapes of the two time series are different in the beginning.
**Step 2: Analyze the middle of the time series**
In the middle of the time series, the blue time series has a more jagged shape with multiple peaks and troughs, while the green time series has a smoother shape with a single peak. The shapes of the two time series are different in the middle.
**Step 3: Analyze the end of the time series**
At the end of the time series, the blue time series decreases rapidly, while the green time series decreases more gradually. The shapes of the two time series are different at the end.
**Conclusion**
Based on the visual inspection of the time series data, it is clear that the blue time series does not have the same shape as the green time series. Therefore, the answer is No.
**Answer**: No}
\end{minipage}
\caption{True positive model response (correct)}
\end{subfigure}
\caption{Model responses for the \emph{InternalBleeding18} dataset, which has dataset context: \emph{This data contains the arterial blood pressure measurements of a pig. The time series is periodic and smooth. In each period, the time series rises to a single peak, then falls back to the baseline.}}
\label{f-InternalBleeding18}
\end{figure}

\begin{figure}
\centering
\begin{subfigure}[b]{\textwidth}
\includegraphics[width=\textwidth]{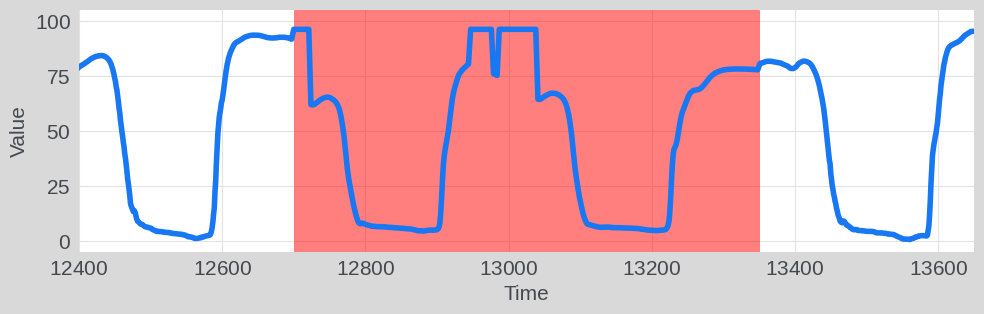}
\caption{Time series values}
\end{subfigure}
\begin{subfigure}[b]{0.48\textwidth}
\includegraphics[width=\textwidth]{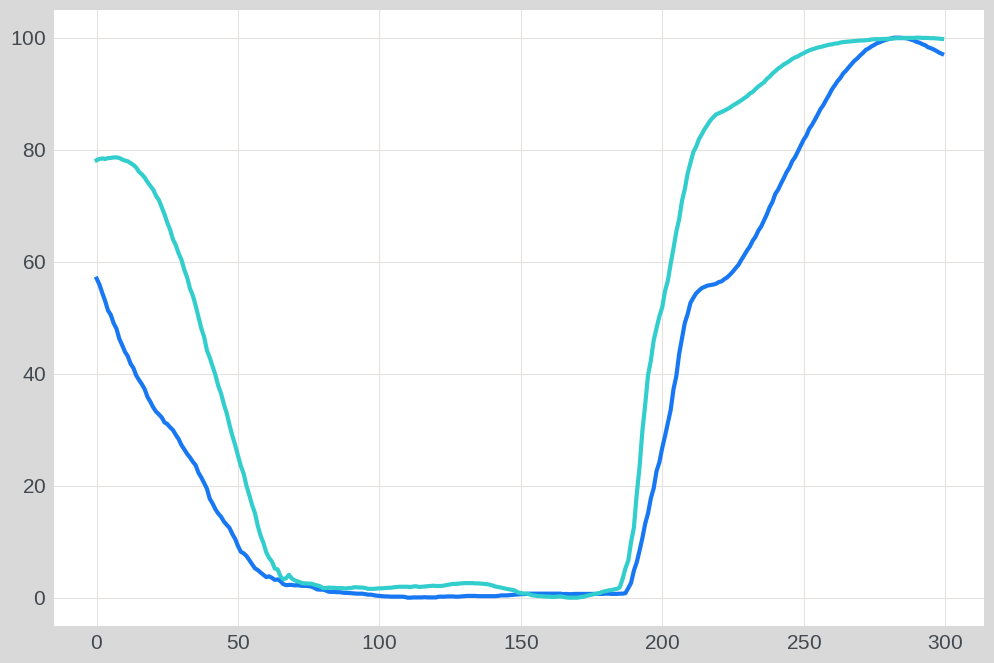}
\caption{False positive example}
\end{subfigure}
\hfill
\begin{subfigure}[b]{0.48\textwidth}
\includegraphics[width=\textwidth]{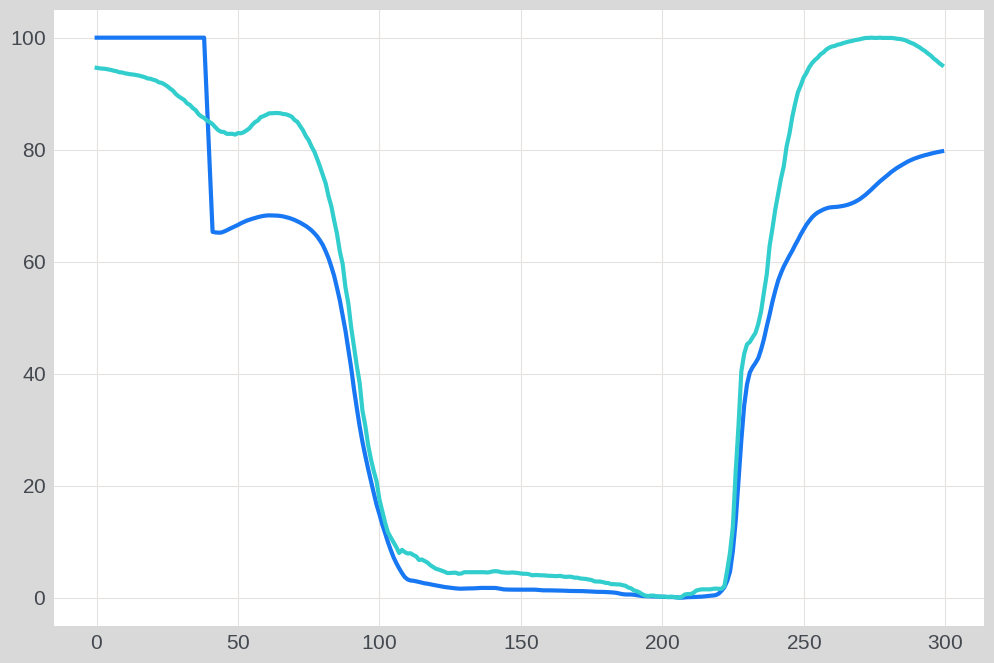}
\caption{True positive example}
\end{subfigure}
\begin{subfigure}[b]{\textwidth}
\vspace{1em}
\begin{minipage}[t]{\textwidth}
\scriptsize
\texttt{**Step 1: Analyze the beginning of the time series**
*   The blue time series starts at approximately 60 units on the y-axis and decreases to around 0 units.
*   The green time series starts at approximately 80 units on the y-axis and decreases to around 0 units.
**Step 2: Analyze the middle of the time series**
*   The blue time series remains at approximately 0 units for a significant portion of the middle section.
*   The green time series also remains at approximately 0 units for a significant portion of the middle section.
**Step 3: Analyze the end of the time series**
*   The blue time series increases from approximately 0 units to around 100 units.
*   The green time series increases from approximately 0 units to around 100 units.
**Conclusion:**
The blue time series has the same shape as the green time series. Both time series exhibit a similar pattern, with an initial decrease, a flat middle section, and a final increase. This suggests that the blue time series also comes from someone walking on a force plate in a biomechanics lab, with periodic, smooth, and consistent data.}
\end{minipage}
\caption{False positive model response (correct)}
\end{subfigure}
\hfill
\begin{subfigure}[b]{\textwidth}
\vspace{1em}
\begin{minipage}[t]{\textwidth}
\scriptsize
\texttt{**Step 1: Analyze the beginning of the time series**
*   The blue time series starts at a high value, around 100.
*   The green time series also starts at a high value, around 90.
**Step 2: Analyze the middle of the time series**
*   The blue time series has a sharp drop in the middle, reaching a low value of around 10.
*   The green time series has a similar sharp drop in the middle, reaching a low value of around 10.
**Step 3: Analyze the end of the time series**
*   The blue time series ends at a high value, around 80.
*   The green time series also ends at a high value, around 90.
**Conclusion**
The blue time series has the same shape as the green time series. Both time series have a similar pattern, with a high starting value, a sharp drop in the middle, and a high ending value. Therefore, the answer is Yes.
**Additional Information**
The data comes from someone walking on a force plate in a biomechanics lab. The data is periodic, with a consistent smooth pattern and shape.}
\end{minipage}
\caption{True positive model response (incorrect)}
\end{subfigure}
\caption{Model responses for the \emph{gaitHunt2} dataset, which has dataset context: \emph{This data comes from someone walking on a force plate in a biomechanics lab. The data is periodic, with a consistent smooth pattern and shape.}}
\label{f-gaitHunt2}
\end{figure}

Figures \ref{f-sddb40}, \ref{f-InternalBleeding18}, and \ref{f-gaitHunt2} show the outputs of \texttt{llama3.2-90b-vision-instruct} on three datasets: \emph{sddb40}, \emph{InternalBleeding18}, and \emph{gaitHunt2}. Figure \ref{f-sddb40} shows an example where the model correctly identified a true positive and a false positive on the \emph{sddb40} heartbeat dataset. Figure \ref{f-InternalBleeding18} shows an example where the model correctly identified a true positive, but failed to identify a false positive, on the \emph{InternalBleeding18} dataset. Figure \ref{f-gaitHunt2} shows an example where the model incorrectly identified a true positive, but correctly identified a false positive on the \emph{gaitHunt2} dataset.

~\newpage~

\begin{table}
\caption{Details about the datasets used. The length of the training and test time series and the length of the single anomaly in the test set are given. The number of true positive (TP) and false positive (FP) intervals found by the $k$-NN baseline anomaly detector.}
\label{t-datasets}
\centering
\begin{tabular}{llllll}
Dataset name & Train len. & Test len. & Anomaly len. & Num. TPs & Num. FPs \\
\midrule
1sddb40                 & 35,000 & 44,795  & 620  & 4 & 1  \\
2sddb40                 & 35,000 & 45,001  & 300  & 2 & 31  \\
CIMIS44AirTemperature5  & 4,000 & 4,184  & 48  & 1 & 0  \\
CIMIS44AirTemperature6  & 4,000 & 4,184  & 48  & 3 & 0  \\
InternalBleeding18      & 2,300 & 5,200  & 102  & 2 & 1  \\
InternalBleeding19      & 3,000 & 4,500  & 10  & 2 & 0  \\
InternalBleeding4       & 1,000 & 6,321  & 358  & 3 & 0  \\
Lab2Cmac011215EPG1      & 5,000 & 25,001  & 50  & 2 & 0  \\
Lab2Cmac011215EPG4      & 6,000 & 24,066  & 130  & 4 & 23  \\
Lab2Cmac011215EPG5      & 7,000 & 22,826  & 130  & 1 & 0  \\
PowerDemand2            & 14,000 & 15,931  & 360  & 3 & 32  \\
TkeepForthMARS          & 3,500 & 7,808  & 97  & 1 & 2  \\
WalkingAceleration5     & 2,700 & 3,984  & 59  & 2 & 0  \\
gaitHunt2               & 18,500 & 45,500  & 650  & 1 & 4  \\
insectEPG1              & 3,000 & 7,001  & 30  & 2 & 6  \\
sddb49                  & 20,000 & 60,000  & 250  & 3 & 0  \\
sel840mECG2             & 20,000 & 37,001  & 370  & 2 & 29  \\
tiltAPB1                & 100,000 & 30,001  & 67  & 1 & 0  \\
\bottomrule
\end{tabular}
\end{table}

\end{document}